\documentstyle[12pt,a4wide,examples]{article}
\newcommand{\scare}[1]{`#1'}
\newcommand{\squote}[1]{`#1'}
\newcommand{\term}[1]{{\sc #1}}
\newcommand{\lingform}[1]{{\it #1}}

\title{Automatic Identification of Support Verbs: \\
A Step Towards a Definition of Semantic Weight}

\author{Mark Dras
\thanks{Reprinted with kind permission from:
"Automatic Identification of Support Verbs: A Step Towards a Definition
of Semantic Weight" in Proceedings of the Eighth Australian Joint
Conference on Artificial Intelligence (World Scientific, Singapore, 1995)
pp 451 - 458.  Copyright by World Scientific Publishing Co. Pte, 1995.}
\\ {\normalsize \it 65 Epping Road, North Ryde NSW 2113, Australia}
\\ {\normalsize Email: \tt t-markdr@microsoft.com}
}

\begin{document}

\thispagestyle{empty}

\date{\vspace{-0.3in}}
\maketitle

\begin{abstract}

Current measures of the readability of texts are very simplistic,
typically based on counts of words or syllables per sentence.  A more
sophisticated analysis needs to take account of the fact that the
particular distributions of meanings across wordings chosen by the
writer, and the consequent variations in syntactic structure, have a
significant effect on readability.

A step towards the required sophistication is provided by the notion
of \term{lexical density} (Halliday, 1985), which suggests that different
words carry different amounts of semantic weight; this idea of semantic
weight is also used implicitly in areas such as information retrieval
and authorship attribution.

Current
definitions of these notions of lexical density and semantic weight are
based on the division of words into
closed and open classes, and on intuition.  This paper develops a
computationally tractable definition of semantic weight,
concentrating on what it means for a word to be semantically light;
the definition involves looking at the frequency of a word in particular
syntactic constructions which are indicative of lightness.
Verbs such as \lingform{make} and \lingform{take}, when they function as
support verbs, are often considered to be semantically light.  To test
our definition, we carried out an experiment based on that of
Grefenstette and Teufel~(1995), where we automatically identify light
instances of these words in a corpus; this was done by incorporating
our frequency-related definition of semantic weight into a statistical
approach similar to that of Grefenstette and Teufel.  The results show
that this is a plausible definition of semantic lightness for verbs,
which can possibly be extended to defining semantic lightness for
other classes of words.

\end{abstract}

\section{Introduction}

There are a number of ways of measuring properties of text, and from
there proceeding to make stylistic judgments; they can be found in style
guides, and include calculating readability indices, counting the number
of passive constructions, and so on.  One attribute of text that
is rarely mentioned explicitly in these style guides, but which underpins
many of the pieces of advice, is that of
\term{semantic density} (see Dras and
Dale, 1995).  Consider the following pair of sentences, taken from
Halliday~(1985):

\begin{examples}
\item \label{ex4}
\begin{subexamples}
\item Slavish imitation of models is nowhere implied.
\item It is not implied anywhere that there are models which should be
slavishly imitated.
\end{subexamples}
\end{examples}

It is apparent that the first of the pair is \scare{denser} than the
second: both express the same (propositional) meaning, but the first does
so in a more compact way.  Halliday terms this \term{lexical density},
\squote{the density with which information is presented} (p68),
and measures it by looking at the proportion of \term{content words}.
Halliday adopts a fairly standard conception of content words as those
which belong to the open word classes:  nouns, verbs, adjectives and so on.
Non-content words are then those that belong to the closed classes,
such as prepositions, auxiliaries and so on.  The non-content,
closed class words are viewed by Halliday as lacking in informational content.

He does, however, note that there are some words on the borderline between
content and non-content words which are lexical items but in many
cases do little more than perform a grammatical function.  These include
the noun \lingform{thing}, as in \lingform{That's a thing I could
do without} (which could be rewritten as \lingform{I could do without
that}), and the verb \lingform {make}, as in \lingform{Christophe
made a decision to come to the Drag Day} (possibly rewritten as
\lingform{Christophe decided to come to the Drag Day}).

In this paper I look at a possible definitional extension of non-content
words, which incorporates the intuition expressed by Halliday that
words like \lingform {make} often contribute little, if any, propositional
meaning to the text.  This new definition is tested by an experiment
modelled on that of Grefenstette and Teufel~(1995), which tries to
find the support verb that particular nominalisations will take---why
\lingform {decision}, for example, takes \lingform {make}, and not
\lingform {have},
\lingform {do}, \lingform {eat} or \lingform {perambulate}.

\section{Lightness of Words}

\subsection{Current views of lightness}

The fact that a word contributes little if any content to a text is used
in a number of areas of linguistics and computational linguistics.
In information retrieval, non-content words are discarded, as they cannot
help to identify the topic of a text.  Mosteller and Wallace~(1984),
on the other hand, retain them and discard the content words when attempting
to statistically determine the authorship of the disputed Federalist
papers, reasoning that while content words may vary across topic,
for a given author non-content words will not.  Halliday~(1985),
as mentioned above, uses them to define the informational density of a
text, in order to compare spoken and written text.

What comprises the class of non-content words is neither uniform nor clearly
defined.  Halliday defines it to be the set of those words which are
part of a closed class system; information retrieval commonly uses a
combination of high-frequency and known function words; Mosteller
and Wallace use a list which was derived
from sources such as the King James Bible.

Halliday proposes that relative frequency of a word can be used to indicate
the amount of information it contributes.  If this is true, the choice of
closed class words to represent non-content words is plausible, since
a given grammatical item (\lingform{the}, \lingform{and}, \lingform{it})
is more likely to have a higher frequency of occurrence than a given
lexical item (\lingform{dog}, \lingform{run}, \lingform{verisimilitude}).
It would also include \lingform{make} and \lingform{thing}, which are
high frequency lexical items.

However, this idea needs to be further refined.  A quick inspection of
a corpus will show that there are a number of words with definite
propositional content
which rank above non-content words in frequency.  In the 8
million word Grolier's Encyclopedia
the verb \lingform{include} (which definitely
conveys information, so it can't be a light constituent) occurs transitively
4284 times, as against \lingform{make}'s 2697 times.

\subsection{A different view}

\lingform{Make} does, however, occur more frequently in constructions
which I will call \term{light constructions}, such as \lingform{make
a decision}; they are mentioned under one name or another by linguists
and style guide authors, and what characterises them is that the light
constituent can be deleted (with some rewriting of the remaining text
to retain grammaticality).  For example, \lingform{make a decision} can
be rewritten as \lingform{decide}, the light element being \lingform{make}.
Jespersen~(1954) is one of the earliest to
note these, commenting on the \term{light verbs} in expressions such
as \lingform{take a walk}.  Style guide writers like
Kane~(1983) mention \scare{deadwood} which can be eliminated
from phrases such as \lingform{It is important for teachers to have
a knowledge of their students} (a possible rewriting being \lingform{It
is important
for teachers to know their students}).  There are quite a few of
these constructions, such as light verbs with noun phrase complements,
light verbs with adjectival complements, and light nouns with post-modifiers
(Dras and Dale, 1995), but this paper only looks at one construction,
the light verbs with NP complements.  By definition, these constructions
will characteristically contain light verbs; this paper therefore
proposes that a modified definition be used for indicating whether
a word can be considered a non-content one: that the word has a high
relative frequency in these light constructions.  In particular, it examines
the relationship of the relative frequency of a verb in these
light verb constructions to its lightness.

It has been suggested that semantic factors are what determine the
relationship between a syntactic construction and its associated light
verb.  Wierzbicka~(1982) proposes a set of semantic rules for determining
the light verb that corresponds to a particular noun object---an
explanation of why
one can \lingform{have a drink} but not \lingform{*have an eat}.  However,
defining these rules by hand for all nouns would be too
time-consuming to be practical.  Grefenstette and Teufel~(1995)
take a statistical approach to finding what is termed the \term{support
verb} for a particular noun.  They look at several nouns, including
\lingform{appeal},  \lingform{proposal}, and \lingform{demand}.  In
a corpus of newspaper articles, they look for occurrences of the noun
and corresponding verb to find the most likely candidate for the support
verb.  They find that the most likely support verb for
\lingform{appeal} is \lingform{make}, which accords with intuition, but for
\lingform{proposal}, their system also finds \lingform{reject}
as an equally likely candidate; and for \lingform {demand}, the most
likely candidate is \lingform {meet}.  In this paper, I conduct a similar
experiment, finding support verbs for given nouns, to test
the definition proposed above: that a word's status regarding content-freeness
is related to its frequency of occurrence in light constructions.

\section{Experiment}

The aim of the experiment is to show that there is a relationship between
relative frequency of verbs in particular constructions and the
content-freeness of these verbs.  A consequence of this is to be able
to choose
the light verb that corresponds to a nominalisation in a
light verb--NP complement
construction---the nominalisation's support verb (SV).
Deverbal nominalisations are chosen as they are the kinds of
grammatical entities which enter into the
SV--NP complement construction.

\subsection{Experiment design}

A way of extracting light verbs from a corpus is to simply take all
verb-object pairs where the object is a deverbal nominalisation.
Grefenstette and Teufel use only \term{local information}, information
that is specific to a particular nominal.
Counting all occurrences of each noun in verb-object pairs yields a
local relative frequency for each verb with respect to that noun.
So, to determine the support
verb for \lingform{proposal} they look only at verbs which co-occur
with the noun \lingform{proposal}.
While it seems intuitively obvious
to native English speakers that \lingform{make} is a more likely
candidate for support verb than \lingform{reject},
the local frequency evidence does not indicate this.
Speakers also use the fact that \lingform{make} is the support
verb for other nominalisations such as \lingform {judgment} and
\lingform{decision}.  I have termed this knowledge \term{global information}.
Counting all occurrences of each verb, regardless of their objects, yields
a global relative frequency for that verb.
In this experiment the local information is combined with the
global information to produce a modified likelihood of being a support
verb.

\subsection{Deriving local and global information}

To gather local and global information, the 1992 version of Grolier's
encyclopedia, tagged by the part-of-speech tagger developed by Brill~(1993),
was used.  A heuristic for producing the local information involved
searching the corpus for the nominal, determining the verb (if any) for which
the nominal was the direct object, and measuring the relative frequency
of these verbs.

The theoretical global information is a measure of how productive a given
support verb is: that is, how many different instances of the SV-NP
construction
it enters into.  The more productive verbs (like \lingform{make})
rank higher on the list than less productive verbs (like \lingform{bear});
this is combined with the local information so that the more
productive verbs, for a particular nominal, are subsequently ranked more
highly than
by the local information alone.  This weighting technique is similar
to that used by Yarowsky~(1992) in the context of sense disambiguation.
In his work he uses counts of words in a window around a key word
to determine the salience of this key word to a particular sense.  These
word counts are weighted so that more common words contribute less;
that is, the less common words are accorded more importance.
We, on
the other hand, want to give more importance to the more common words,
given our assumption that it is high relative frequency in particular
constructions that helps define semantic lightness.

Grefenstette and Teufel note that a confounding factor in the local
information, when picking out nominals and their associated verbs, is that
the nominal may have become \term{concretised}.
Generally, nominals represent an abstract concept, being essentially events
represented in noun form; but it is possible for the nominal to represent
a physical embodiment of that concept.
For example:

\begin{examples}
\item \label{ex6}
\begin{subexamples}
\item He made his formal proposal to the full committee.
\item He put the proposal in the drawer.
\end{subexamples}
\end{examples}

The abstract and concretised versions will tend to have different associated
verbs.  However, if the relative frequency hypothesis is true,
and the global information is an accurate reflection of the innate lightness
of a verb, this will elevate the light verb over the \scare{heavier} ones,
which will be more likely to be associated with the concretised forms.

In practice, the global information is calculated from the aggregate of
the local data.  This means that there is a lot of noise---all
of the incorrect candidates for support verb are included in the global
information---but again, if the relative frequency hypothesis is
true, the relative frequency of the support verb in the local information
will be high (although not necessarily the highest), while this is
not true for non-support verbs.  So aggregating all of these should reinforce
the support verbs and not the others.

\subsection{Generating nominalisations}

To construct the global information, a comprehensive list of
deverbal nominalisations
is needed, together with the associated support verbs, determined
from the local information.  To generate this list in a partially
automated manner, Longman's
Dictionary of Contemporary English (LDOCE) was used, including both built-in
information and a heuristic: a nominal is an event represented in noun form,
so the procedure used here for deriving a list of them involved looking
for nouns with associated \term{stem verbs}.

Some verbs have this information encoded in their entries: for example,
\lingform{adjust} lists \lingform{adjustment} as its nominalisation;
there were 257 verbs in this category.  For others,
an automatic orthographic heuristic that matched nouns with verbs was
manually filtered to produce 1414 more deverbal nominalisations.
A system to identify support verbs for nominalisations was implemented
by tabulating all the verbs for which these nominals
were the direct object.

The list of nominals did not
cover some of
the nominals from the test set (listed in the table below).
The local information was generated for
each of the excluded test set nominals and aggregated into the global
information.
Candidates for support verb were ranked in order of the product
of local and global relative frequency of each candidate verb.

\subsection{The test set and results}

The light verb constructions and their constituent nominals used for testing
were taken from a range of sources, so that they would not be biased
to one particular genre.

\begin{table*}
\begin{tabular} {|l|l|c|l|l|l|}
\hline
Source Text & Verb & Reference & Choice 1 & Choice 2 & Ratio \\
\hline
\hline
make an attempt & attempt & [\ref{DD}] & make & include & 9.36 \\
\hline
make a change & change & [\ref{DD}] & make & produce & 1.85 \\
\hline
make a concession & concede & [\ref{DD}] & make & include & 11.47 \\
\hline
make a demand & demand & [\ref{GT}] & make & create & 1.03 \\
\hline
make a distinction & distinguish & [\ref{DD}] & make & have & 3.04 \\
\hline
have a drink (of) & drink & [\ref{W}] & become & N/A & N/A \\
\hline
have an effect (on) & affect & [\ref{DD}] & have & produce & 3.04 \\
\hline
have a feeling & feel & [\ref{Ha}] & have & produce & 3.27 \\
\hline
make a gift (of) & give & [\ref{Ha}] & have & include & 9.89 \\
\hline
do harm (to) & harm & [\ref{Hu}] & cause & do & 1.26 \\
\hline
make a judgment & judge & [\ref{DD}] & make & have & 2.43 \\
\hline
have a knowledge (of) & know & [\ref{K}] & have & use & 12.36 \\
\hline
make progress & progress & [\ref{Ha}] & make & allow & 64.33 \\
\hline
make a proposal & propose & [\ref{GT}] & make & include & 1.10 \\
\hline
bear a resemblance (to) & resemble & [\ref{Hu}] & bear & have & 2.64 \\
\hline
give a shove (to) & shove & [\ref{Ha}] & N/A & N/A & N/A \\
\hline
have a snooze & snooze & [\ref{Ha}] & N/A & N/A & N/A \\
\hline
make use (of) & use & [\ref{DD}] & make & have & 6.55 \\
\hline
\end{tabular}
\caption{Support verb candidates} \label{table1}
\end{table*}

\vspace{0.15in}

The test set and results are summarised
in Table~\ref{table1}; the table contains:

\begin{itemize}
\item the source text;
\item the corresponding verb, which the source can be rewritten as;
\item the reference for the source text;
\item the system's first choice candidate for support verb for the source
text's constituent nominalisation;
\item the system's second choice; and
\item the ratio of the adjusted frequency, which is defined as the product
of local
and global relative frequencies, for the first and second choices.
\end{itemize}

\subsection{Discussion}

Of the 18 examples, 13 of the choices for support verb match the corresponding
one from the source text.  Of the five incorrect ones, three
were incorrect because of lack of data: there are no occurrences of
\lingform{snooze}
or \lingform{shove} as direct objects of verbs in Grolier's, most
probably because they belong to a more informal register than that
used in encyclopedias.  Similarly, \lingform{have a drink} is an informal
phrase that would not normally be found in an encyclopedia.

Another of the incorrect cases, \lingform{harm}, had \lingform{cause}
as the proposed alternative.  This is an equally valid support verb,
and in any case, \lingform{do} was the second choice by only a small
margin.  This is true for a number of cases: where there is an alternative
support verb to the one used in the source text, the second alternative
represents another plausible choice, and the frequency ratio margin
is small (for example, for \lingform{change} and \lingform{resemblance}).

So if only the cases where enough data exists are considered, and if
alternative support verbs are allowed, the success rate becomes 14 of 15.

\section{Conclusion}

It is apparent that what constitutes a valid light verb construction depends
on the genre and register of the text.  More accurate results could
no doubt be obtained by using a corpus that was more representative
of general English.  Also, where shortcuts were taken (for example,
by not removing concretised nominalisations), more precision could be
obtained.  Notwithstanding these considerations, using the relative
frequency of a verb in light constructions seems to be a fairly good
indicator of a verb's content-freeness, providing plausible choices for
support
verbs for nominalisations.

Further work will involve extending this definition to other light
constructions---light verbs with adjectival complements, and light nouns
with post-modifiers---and fitting closed-class words, which traditionally
comprise the class of non-content words, into the framework.  This framework
can then be used to give a more accurate indication of lexical density; to
more accurately choose words to leave out, or include in, in the fields of
information retrieval and authorship attribution; and to fine-tune
stylistic judgments.

\section{Acknowledgements}
This work has benefited from the kind assistance of Robert Dale and
Mark Lauer.  I'd also like to thank Mike Johnson for many fruitful
discussions.  Financial support is gratefully acknowledged from
the Australian Government and the Microsoft Institute.

\end{document}